\documentclass[letterpaper]{article} 
\usepackage{aaai2026}  
\usepackage{times}  
\usepackage{helvet}  
\usepackage{courier}  
\usepackage[hyphens]{url}  
\usepackage{graphicx} 
\usepackage{natbib}  
\usepackage{caption} 
\usepackage{algorithm}
\usepackage{algorithmic}
\usepackage{tabularx}
\usepackage{amsmath}
\usepackage{amsfonts}
\usepackage{symbols}
\usepackage{booktabs}    
\usepackage{multirow}
\usepackage{placeins}
\usepackage[utf8]{inputenc}

\usepackage{newfloat}
\usepackage{listings}
\DeclareCaptionStyle{ruled}{labelfont=normalfont,labelsep=colon,strut=off} 
\floatstyle{ruled}
\newfloat{listing}{tb}{lst}{}
\floatname{listing}{Listing}
\nocopyright

\usepackage{booktabs}
\usepackage{multirow}
\usepackage{amssymb, pifont}
\usepackage{graphicx}

\usepackage{amsmath}
\usepackage{multirow}
\usepackage{amssymb}
\usepackage{pifont}
\usepackage[export]{adjustbox}
\usepackage{subcaption}
\usepackage{booktabs}
\usepackage{makecell}
\usepackage{enumitem}
\usepackage{arydshln}
\usepackage{pifont}
\newcommand{\xmark}{\ding{55}}

\title{SignMouth: Leveraging Mouthing Cues for Sign Language Translation by Multimodal Contrastive Fusion}
\author {
    Wenfang Wu\textsuperscript{\rm 1,2},
    Tingting Yuan\textsuperscript{\rm 3},
    Yupeng Li\textsuperscript{\rm 4},
    Daling Wang\textsuperscript{\rm 2},
    Xiaoming Fu\textsuperscript{\rm 1}
}
\affiliations {
   \textsuperscript{\rm 1}Georg-August Universität Göttingen, Germany\\
    \textsuperscript{\rm 2}Northeastern University, Shenyang, China\\
    \textsuperscript{\rm 2}Fachhochschulstudiengänge Krems IMC, Austria\\
    \textsuperscript{\rm 3}Hong Kong Baptist University, Hong Kong, China\\
}

\begin{document}

\maketitle

\begin{abstract}
Sign language translation (SLT) aims to translate natural language from sign language videos, serving as a vital bridge for inclusive communication. 
While recent advances leverage powerful visual backbones and large language models, most approaches mainly focus on manual signals (hand gestures) and tend to overlook non-manual cues like mouthing. 
In fact, mouthing conveys essential linguistic information in sign languages and plays a crucial role in disambiguating visually similar signs. 
In this paper, we propose SignMouth, a novel framework to improve the accuracy of sign language translation. 
It fuses manual and non-manual cues, specifically spatial gesture and mouthing features. 
Besides, SignMouth introduces a hierarchical contrastive learning framework with multi-level alignment objectives, ensuring semantic consistency across spatial-mouthing and visual-text modalities. 
Extensive experiments on two benchmark datasets, PHOENIX14T and How2Sign, demonstrate the superiority of our approach. 
For example, on PHOENIX14T, in the Gloss-free setting, SignMouth surpasses the previous state-of-the-art model SpaMo, improving BLEU-4 from 24.32 to 24.71, and ROUGE from 46.57 to 48.38.\footnote{The code will be made public after the paper is published.}
\end{abstract}


\section{Introduction}
Sign languages~\cite{nunez2023survey, rastgoo2024survey} are fully developed visual languages conveyed through hand gestures, body movements, and facial expressions. 
They serve as the primary means of communication for individuals who are Deaf, hard of hearing, or unable to speak. 
Translating sign language into natural language, a task known as Sign Language Translation (SLT)~\cite{camgoz2018neural}, is essential for promoting accessibility and facilitating communication between signers and non-signers. 
SLT presents a complex cross-modal challenge that involves aligning high-dimensional visual inputs with natural language outputs. 



\begin{figure}[!t]
  \centering
  \includegraphics[width=0.7\linewidth]{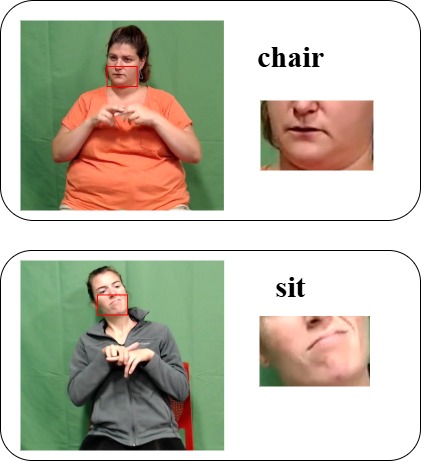} 
\caption{The sample of the similar signs in SLT.}
\label{fig:sample}
\end{figure}

With the advancement of Large Language Models (LLMs), visual signals can now be more effectively translated into natural language. 
This progress has enabled significant improvements in SLT~\cite{wong2024sign2gpt, hwang-etal-2025-efficient, gong2024llms}, making it possible to generate fluent and accurate spoken or written sentences directly from sign language videos.
However, existing works primarily rely on manual cues (i.e., hand gestures), which are often visually similar and challenging to differentiate.
For example, as illustrated in Figure~\ref{fig:sample}, the signs for ``chair'' and ``sit'' share the same hand configuration but differ in their accompanying mouth shapes. 
Incorporating non-manual signals, such as mouthing, is crucial for effective disambiguation and improved translation performance.
Beyond disambiguation, mouthing also conveys important syntactic and semantic information. 
It can indicate grammatical properties such as tense, modality, and emphasis, as well as encode function words and particles, which are particularly challenging to capture in gloss-free translation settings where no intermediate annotations are provided.


Integrating mouthing information into SLT systems poses several unique challenges. 
First, mouthing and hand gestures differ significantly in their spatial focus. 
While hand gestures involve broad body movements, mouthing is confined to subtle changes in the lower face region.
Without appropriate preprocessing to isolate the mouth region, raw video inputs may introduce substantial noise, making it difficult for the model to extract useful mouthing features. Second, when multiple visual streams (hand features and lip movements) are combined, effective cross-modal alignment becomes essential. Misalignment between modalities can degrade translation performance, especially in gloss-free SLT settings where no intermediate supervision is available to guide the fusion process.

To address these challenges, we propose SignMouth, a novel multimodal contrastive fusion framework for SLT that jointly captures manual signals, such as hand gestures, and non-manual cues, particularly mouthing.
Our approach adopts a dual-stream architecture that independently encodes the visual input of the full frame and the localized mouth region, followed by a flexible fusion module with gated mechanisms to integrate the two modalities. 
To encourage effective cross-modal interaction, we introduce two contrastive learning objectives: one enforces consistency between hand gesture and mouthing representations, while the other aligns visual features with corresponding textual semantics. 
This multi-level supervision enhances modality alignment and enables the model to better disambiguate visually similar signs, ultimately improving translation quality in gloss-free settings.
We conduct extensive experiments on two benchmark datasets: PHOENIX14T and How2Sign. 
Our model consistently outperforms existing SLT baselines, particularly in scenarios where mouthing plays a key disambiguation role. 
Moreover, ablation studies demonstrate the effectiveness of mouthing integration and contrastive objectives.

\noindent{Our main contributions are summarized as follows:}
\begin{itemize}
\item We propose SignMouth, a multimodal contrastive fusion framework for SLT that jointly models manual (hand gestures) and non-manual (mouthing) cues, effectively addressing sign ambiguity in gloss-free translation settings.
\item We design a dual-stream architecture with gated fusion and introduce multi-level contrastive objectives to align hand gestures with mouthing and visual features with target text, enhancing cross-modal representation learning.
\item Our approach achieves consistent improvements on PHOENIX-2014T (+0.39 BLEU-4) and How2Sign (+0.64 BLEU-4) over strong baselines, validating the effectiveness of mouthing integration and contrastive learning.
\end{itemize}

\section{Related Work}

\subsection{Gloss-based Sign Language Translation}
Gloss-based methods decompose the SLT task into two stages: Sign Language Recognition (SLR), which converts sign videos into gloss sequences (a structured written representation of sign units), followed by Neural Machine Translation (NMT) to transform glosses into natural language text~\cite{bohavcek2022sign, chen2022simple, zhou2021improving, chen2022two}. 
Early work such as \citet{camgoz2018neural} adopted this framework, achieving strong results on benchmarks like PHOENIX-2014T. 
Later advances incorporated powerful language models MMTLB~\cite{chen2022simple}, multi-stream visual encoders TS-SLT~\cite{chen2022two}, and joint sign-related task training frameworks SLTUNet~\cite{zhang2023sltunet}. 
However, gloss-based methods depend on costly, language-specific gloss annotations, limiting their scalability and generalization in real-world applications.

\subsection{Gloss-free Sign Language Translation}
To overcome the reliance on costly gloss annotations in traditional SLT pipelines, gloss-free approaches have emerged that directly translate sign language videos into spoken language without intermediate supervision. 
GFSLT-VLP~\cite{zhou2023gloss} aligns sign video features with spoken sentences through visual-linguistic pretraining, improving compatibility with LLMs. 
Sign2GPT~\cite{wong2024sign2gpt} further enriches representation learning by aligning visual features with pseudo-glosses and POS-tagged words. 
SpaMo~\cite{hwang-etal-2025-efficient} decouples spatial and motion features using off-the-shelf visual encoders and feeds them into LLMs with language prompts. These advances highlight the potential of gloss-free SLT when combined with multimodal fusion and LLM-based reasoning. 

Despite recent progress, SLT remains challenging due to the complexity of sign languages. 
Most existing methods rely solely on manual cues (i.e., hand gestures). These cues are important, but they are often visually similar and insufficient for disambiguation.
Building on prior analysis \cite{wadhawan2021sign,7406418}, we observe that facial information, particularly mouthing, conveys important linguistic cues for sign language understanding. 
However, such non-manual features are often overlooked in existing SLT approaches.

\section{Method}
\begin{figure*}[t]
    \centering
    \includegraphics[width=0.95\textwidth]{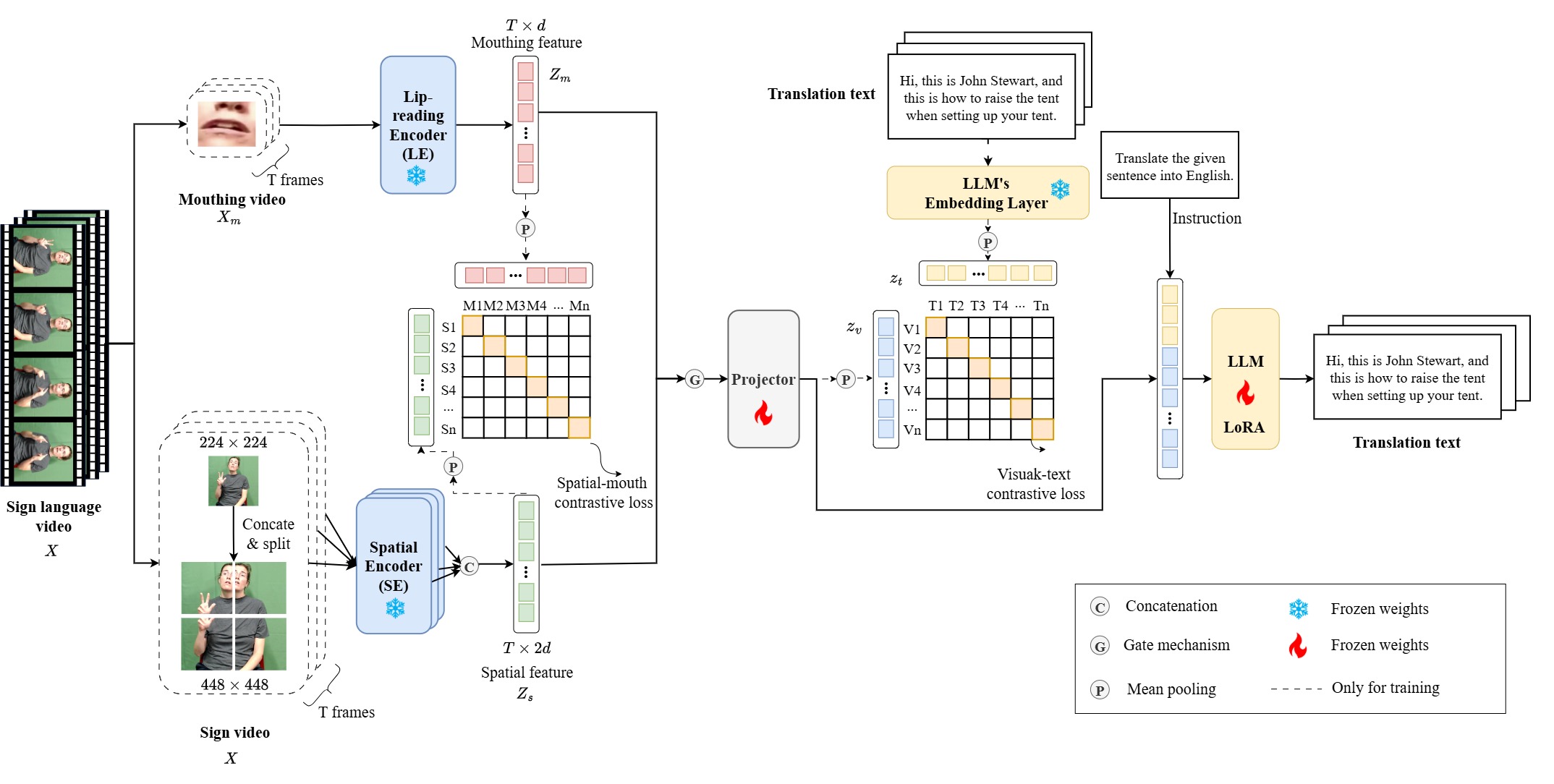}
    \caption{The SignMouth framework consists of four main components: (1) a spatial encoder (SE) based on Vision Transformer (ViT) to capture spatial representations of hand gestures; (2) a lip-reading encoder (LE) to model mouthing shape dynamics that complement gesture semantics; (3) a multimodal contrastive fusion module that introduces a hierarchical contrastive learning strategy to align and integrate gesture, mouthing, and textual features; and (4) a LLM that takes the fused visual features along with language-instructive prompts and performs sign-to-text translation, fine-tuned with Low-Rank Adaptation (LoRA). C represents concatenate; G is gate mechanism; P is mean pooling.} 
    \label{fig:overview}
\end{figure*}

\subsection{Overview}\label{sec3.1}
Given a sign language video $X = \{x_i\}_{i=1}^T$, where each frame $x_i \in \mathbb{R}^{H \times W}$ represents a single RGB frame, the objective of Sign Language Translation (SLT) is to generate a natural language sentence $Y = \{y_j\}_{j=1}^U$, consisting of $U$ words. 
Prior gloss-free SLT approaches~\cite{zhou2023gloss, wong2024sign2gpt, hwang-etal-2025-efficient} have focused on extracting visual representations from signers’ gestures and feeding them into sequence-to-sequence architectures. 
However, these methods often rely solely on manual signals (hand gestures) and overlook the rich linguistic cues provided by non-manual features, particularly mouthing.

As illustrated in Figure~\ref{fig:overview}, our proposed framework introduces a dual-stream encoder that processes both sign gesture cues and mouthing information. 
First, a spatial encoder extracts fine-grained manual features $Z_s$ from each cropped sign frame. 
Simultaneously, we isolate the mouth region using facial landmark detection and construct a mouthing video, from which a frozen lip-reading encoder derives the non-manual feature stream $Z_m$.

To effectively combine both modalities, we introduce a flexible fusion module, which integrates $Z_s$ and $Z_m$ via gated fusion, resulting in a unified representation $Z_{fused}$. 
This fused representation is projected into the embedding space of the LLM (Flan-T5), guided by a task-specific prompt.

To further strengthen the interaction between modalities and improve generalization, we incorporate multi-level contrastive learning objectives: one encourages alignment between visual and mouthing features to disambiguate visually similar gestures; the other aligns sign features with textual embeddings from the LLM to ensure semantic consistency.
This training scheme promotes robust multimodal integration and enables the model to generate more accurate and fluent translations, even in challenging gloss-free settings.

\subsection{Gesture Feature Extraction}\label{sec3.2}
Following SpaMo~\cite{hwang-etal-2025-efficient}, we aim to capture the spatial configuration of each sign frame, such as handshape and position, using a frozen image encoder (e.g., CLIP ViT-L/14) as our Spatial Encoder (SE).
To enhance spatial representations without fine-tuning the pre-trained backbone, we adopt \textit{Scaling on Scales (S$^2$)}, a parameter-free multi-scale feature enrichment strategy.

Specifically, each sign frame $x_i \in \mathbb{R}^{H \times W \times 3}$ is resized to multiple resolutions, including $224 \times 224$ and $448 \times 448$. For the higher resolution input (e.g., $448 \times 448$), we divide it into four non-overlapping $224 \times 224$ patches. Each patch, along with the original $224 \times 224$ image, is independently encoded by the frozen image encoder to obtain patch-level embeddings. These embeddings are then aggregated, typically via mean pooling, and concatenated with the original frame embedding to yield a multi-scale spatial feature:

\[
z_s^{\text{frame}} = \text{concat}\left(f(x_i^{224}), \text{pool}\left(f(x_{i,1}^{448}), \ldots, f(x_{i,K}^{448})\right)\right),
\]
where $f(\cdot)$ denotes the frozen image encoder and $K$ is the number of sub-patches (e.g., $K=4$).

The spatial embeddings across the entire video sequence $X = \{x_i\}_{i=1}^T$ form a temporal sequence:
\[
Z_s = \{z_s^{\text{frame1}}, z_s^{\text{frame2}}, \ldots, z_s^{\text{frameT}}\} \in \mathbb{R}^{T \times 2d},
\]
where $2d$ is the dimensionality of the concatenated features. This design allows the SE to jointly capture fine-grained local details and global spatial context.

\subsection{Mouthing Feature Extraction}\label{sec3.3}
Lip-reading Encoder (LE) derives phonological features from the original sign video \( X \) by focusing on the mouth region. 
Similar to the SE, we employ a pre-trained model Av-HuBERT~\cite{shi2022learning}, which remains frozen during training to ensure stable and robust feature extraction. 

\begin{figure}[!t]
  \centering
  \includegraphics[width=0.4\linewidth]{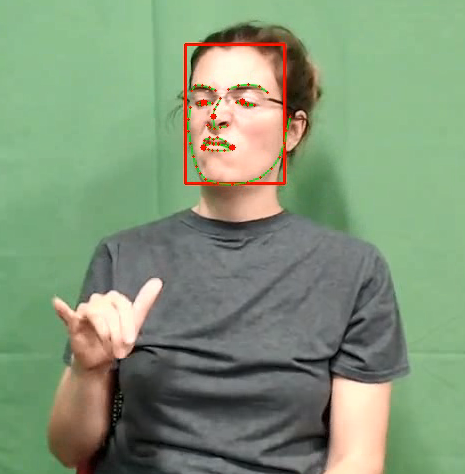} 
\caption{The sample of detected face region.}
\label{fig:face}
\end{figure}

To isolate the mouth region, we use the IBUG face alignment toolkit, which employs a 2D facial landmark detector based on the FAN architecture~\cite{bulat2017far}. Given a full-frame video, the detector predicts 68 facial landmarks. 
We extract the subset corresponding to the mouth region and compute a bounding box to crop the mouth area, shown in Figure~\ref{fig:face}. 
The resulting mouth frames are spatially aligned and temporally stacked to construct the mouthing sequence, denoted as \( X_m \).

The mouthing sequence \( X_m \) is then passed through LE to extract phonological representations:
\[
Z_m = \{ z_m^{\text{frame}1}, z_m^{\text{frame}2}, \ldots, z_m^{\text{frame}T} \} \in \mathbb{R}^{T \times d},
\]
where \( T \) is the number of frames and \( d \) is the feature dimension. 
These features capture fine-grained articulation patterns and lip shapes, providing complementary non-manual cues to the manual sign features extracted by SE.

\subsection{Multimodal Contrastive Fusion}\label{sec3.4}
To effectively integrate complementary information from manual (hand gesture) and non-manual (mouthing) cues, we design a Multimodal Contrastive Fusion (MCF) module that leverages gated fusion followed by temporal modeling.

\paragraph{Gated Fusion.}
Let $Z_s \in \mathbb{R}^{T \times 2d}$ and $Z_m \in \mathbb{R}^{T \times d}$ denote the gesture and mouthing features extracted from the spatial encoder and lip-reading encoder, respectively. 
We first project \( Z_s \) to a lower-dimensional representation
\( Z_s' \in \mathbb{R}^{T \times d} \) via a linear mapping. We then concatenate \( Z_s' \) and \( Z_m \), and compute a gating vector:
\[
\mathbf{g} = \sigma(\text{MLP}([Z_s'; Z_m])),
\]
where \( \sigma \) denotes the sigmoid function. The fused representation is:
\[
Z_{\text{fused}} = \mathbf{g} \odot Z_s' + (1 - \mathbf{g}) \odot Z_m.
\]

\paragraph{Temporal Modeling.}
To capture temporal dependencies, we feed $Z_{fused}$ into a 1D convolutional encoder (Projector in Figure~\ref{fig:overview}):
\[
Z_{conv} = \text{Conv1D}(Z_{fused}^T),
\]
where the input is transposed to shape $[d, T]$ for convolution. The output is then transposed back to $[T, d]$ and used as the visual input to the language model.

\paragraph{Contrastive Supervision.}
To encourage effective modality alignment, we introduce two contrastive objectives:

\begin{itemize}
    \item \textbf{Visual-Text Contrastive Loss} ($\mathcal{L}_{\text{vt}}$): This loss aligns the video-level visual representation with its corresponding textual description. Given a batch of $N$ video-text pairs, we obtain a visual summary embedding $z_v \in \mathbb{R}^d$ (e.g., via mean pooling over $Z_{conv}$) and a sentence embedding $z_t \in \mathbb{R}^d$ (from the Flan-T5 encoder). We adopt an InfoNCE-style contrastive loss:
    \[
    \mathcal{L}_{\text{vt}} = -\frac{1}{N} \sum_{i=1}^{N} \log \frac{
    \exp(\mathrm{sim}(\mathbf{z}_v^i, \mathbf{z}_t^i)/\tau)}{
    \sum_{j=1}^{N} \exp(\mathrm{sim}(\mathbf{z}_v^i, \mathbf{z}_t^j)/\tau)},
    \]
    where $\mathrm{sim}(\cdot, \cdot)$ denotes cosine similarity, and $\tau$ is a temperature hyperparameter.
    
    \item \textbf{Spatial-mouthing Contrastive Loss} ($\mathcal{L}_{\text{sm}}$): This loss encourages consistency between spatial and mouthing modalities at the video level. 
    We summarize the gesture sequence $Z_s$ and mouthing sequence $Z_m$ into global embeddings $z_s, z_m \in \mathbb{R}^d$ (e.g., via temporal pooling and projection). The loss is:
    \[
    \mathcal{L}_{\text{sm}} = -\frac{1}{N} \sum_{i=1}^{N} \log \frac{
    \exp(\mathrm{sim}(\mathbf{z}_s^i, \mathbf{z}_m^i)/\tau)}{
    \sum_{j=1}^{N} \exp(\mathrm{sim}(\mathbf{z}_s^i, \mathbf{z}_m^j)/\tau)},
    \]
    aligning spatial and mouthing features from the same video while pushing apart mismatched pairs.
\end{itemize}

\subsection{Training details}\label{sec3.5}
Our model is trained in an end-to-end fashion with joint supervision from the translation and contrastive objectives. The total loss is defined as:
\[
\mathcal{L}_{\text{total}} = \mathcal{L}_{\text{trans}} + \alpha \mathcal{L}_{\text{vt}} + \beta \mathcal{L}_{\text{sm}},
\]
where $\mathcal{L}_{\text{trans}}$ is the cross-entropy loss for sign-to-text translation, $\mathcal{L}_{\text{vt}}$ denotes the visual-text contrastive loss, and $\mathcal{L}_{\text{sm}}$ represents the contrastive alignment between spatial and mouthing features. 
The weights $\alpha$ and $\beta$ control the importance of each auxiliary objective.

We utilize a pre-trained LLM Flan-T5~\cite{chung2024scaling} as the decoder. To efficiently adapt the LLM to the SLT task, we apply Low-Rank Adaptation (LoRA)~\cite{hu2022lora} to both the self-attention and feedforward layers. This approach allows us to fine-tune only a small number of parameters, significantly reducing memory and computation overhead.

\section{Experiment}

\begin{table*}[ht]
    \centering
    \scriptsize
    \resizebox{\linewidth}{!}{%
        \begin{tabular}{l c c c c c c c c l} 
        \toprule 
         \textbf{Dataset} &\textbf{Language} &\textbf{Video} & \textbf{Gloss} &\textbf{Vocab} & \textbf{Duration (h)} & \textbf{Samples} &\textbf{Train} / \textbf{Valid} / \textbf{Test}& \textbf{Signers}& \textbf{Domain}\\ \midrule
         PHOENIX14T~\cite{camgoz2018neural}& German&video& \(\checkmark\)&2,887&10.96&8257& 7,096 / 519 / 642& 9 & Weather \\
         How2Sign~\cite{duarte2021how2sign} & American&Multi-view video& \xmark &15,686&80&35,000& 31,128 / 1,741 / 2,322& 11&Instructional \\ \bottomrule
        \end{tabular}
    }
    \caption{Statistics of two sign language datasets used in this work.} 
    \label{tab:dataset}
\end{table*}

\subsection{Implementation Details}
We train our model on 2 NVIDIA A100 GPUs with a total batch size of 16. Flan-T5-XL is used as the language backbone and fine-tuned using LoRA for efficient adaptation. 
Spatial gesture and mouth-region features are projected into a shared embedding space and fused via a gated mechanism, followed by a 1D convolutional layer to capture temporal dynamics. 
The model is trained for 50 epochs without warm-up, using the AdamW optimizer with learning rates of 5e-5 for visual modules and 1e-4 for the language model. Mouth-region video clips are extracted using FAN-based 2D facial landmark detection.
For the multimodal contrastive fusion part, we empirically set the weight $\alpha=1.0$ to emphasize alignment between visual features and textual semantics, and $\beta=0.2$ to encourage auxiliary consistency between hand gestures and mouthing.
\subsection{Datasets and Metrics}

\paragraph{Dataset}
We evaluate our approach on two publicly available SLT datasets that vary in language, domain, and complexity (see details in Table~\ref{tab:dataset}): PHOENIX14T~\cite{camgoz2018neural} and How2Sign~\cite{duarte2021how2sign}. 
PHOENIX14T is a benchmark dataset for German Sign Language (DGS) translation, collected from weather forecast broadcasts. It represents a closed-domain setting and contains 8,257 sign language video clips. Each clip has an average of 116 frames, and the dataset provides aligned spoken German translations, making it suitable for evaluating sentence-level SLT models.
How2Sign is a large-scale dataset for American Sign Language (ASL), covering a more diverse and open-domain instructional context, including cooking, crafting, and general tutorials. It contains over 13,000 sign language video clips captured from multiple camera views, with an average length of 173 frames per clip. This dataset poses greater challenges for SLT due to the wider range of vocabulary and variability in signing styles. 
As detailed in the Appendix, the dataset statistics include the distribution of different sample types, the size of the dataset, and other relevant information.

\paragraph{Evaluation metrics}
We assess the quality of sign language translation using a suite of widely adopted natural language generation metrics: BLEU-1 to BLEU-4~\cite{post2018call} and ROUGE-L~\cite{lin2004automatic}.
BLEU-1 to BLEU-4 evaluate n-gram precision between the predicted and reference sentences. 
We use the SacreBLEU implementation with 13a tokenization to ensure fair and reproducible comparison. 
ROUGE-L computes the longest common subsequence (LCS) between candidate and reference texts, focusing on sequence-level matching and recall.

\subsection{Baselines}

\paragraph{Gloss-Based SLT Methods.}

MMTLB~\cite{chen2022simple} is a modular framework that first performs Sign Language Recognition (SLR) into glosses and then applies Neural Machine Translation (NMT).
BN-TIN-Transf.+SignBT~\cite{zhou2021improving} enhances SLT performance by introducing a sign back-translation strategy that synthesizes sign-text pairs from large-scale spoken texts via a text-to-gloss and gloss-to-sign pipeline.
SLTU$_{Net}$~\cite{zhang2023sltunet} is a unified network jointly trained for SLR and NMT, leveraging multi-task learning to enhance gloss-based translation.
TwoStream-SLT~\cite{chen2022two} is a two-stream model that fuses RGB and keypoint features for robust gloss prediction.

\paragraph{Gloss-Free SLT Methods.}
These gloss-free SLT methods focus on different aspects of sign language translation. 
OpenSLT~\cite{tarres2023sign} and YT-ASL-SLT~\cite{uthus2024youtube} introduced new datasets or benchmarks for How2Sign and YouTube-ASL, which have helped standardize SLT research. 
Methods like GFSLT-VLP~\cite{zhou2023gloss}, FLa-LLM~\cite{chen2024factorized}, and SpaMo~\cite{hwang-etal-2025-efficient} concentrate on extracting gesture features, including spatial and motion characteristics, to improve sign language understanding and translation. 
Sign2GPT~\cite{wong2024sign2gpt}, SignLLM~\cite{gong2024llms}, and GloFE-VN~\cite{lin2023gloss} focus on generating intermediate representations similar to gloss, bridging the gap between visual features and textual translation. 
Lastly, SSVP-SLT~\cite{uthus2024youtube} highlights the importance of privacy protection, addressing how to balance effective translation with data privacy concerns. 

These baselines represent the state of the art in SLT across multiple paradigms. Our method distinguishes itself by introducing multi-level contrastive fusion of hand gestures and mouthing, improving translation quality, especially in gloss-free scenarios.

\subsection{Main Results}
\paragraph{Results on PHOENIX14T}
We evaluate our proposed method, SignMouth, on the PHOENIX14T dataset and compare it against both gloss-based and gloss-free sign language translation baselines (Table~\ref{tab:res_p14t_csl}). 
In gloss-free setting, SignMouth achieves better performance with a BLEU-4 score of 24.71, outperforming recent strong baselines such as SpaMo~\cite{hwang-etal-2025-efficient} (24.32) and Sign2GPT~\cite{wong2024sign2gpt} (22.52), without requiring any visual feature fine-tuning. 
SignMouth also achieves the highest BLEU-1 (50.37) and BLEU-3 (29.94) scores among all gloss-free methods, as well as a competitive ROUGE score of 48.38, second only to Sign2GPT. These results clearly demonstrate the effectiveness of introducing mouthing features, which provide critical disambiguation cues for visually similar signs and significantly improve translation performance in the gloss-free setting.

\paragraph{Results on How2Sign}
To further validate the generalization ability of our model, we evaluate SignMouth on the How2Sign dataset and compare it against recent gloss-free and weakly gloss-free baselines (Table~\ref{tab:res_h2s}). SignMouth achieves new state-of-the-art results across all metrics in the gloss-free setting, with a BLEU-4 score of 10.75, outperforming the best previous model SpaMo~\cite{hwang-etal-2025-efficient} by +0.64 BLEU-4. In addition, SignMouth reaches the highest BLEU-1 (33.88) and ROUGE (31.44) scores, further highlighting its superior translation quality.

\begin{table*}[ht]
    \centering{
        \begin{tabular}{c l c ccccc} \toprule
        \textbf{Setting}& \textbf{Methods} & \textbf{Vis. Fi}& \textbf{B1} &\textbf{B2} &\textbf{B3} &\textbf{B4} &\textbf{RL} \\ \midrule
        \multirow{4}{*}{Gloss-based} 
        & BN-TIN-Transf.+SignBT~\cite{zhou2021improving}& \xmark & 50.80 &37.75 &29.72 &24.32 &49.54 \\
        & MMTLB~\cite{chen2022simple} & \checkmark & 53.97 &41.75 &33.84 &28.39 &52.65 \\
        & TwoStream-SLT~\cite{chen2022two} & \checkmark & 54.90 &42.43 &34.46 &28.95 &53.48 \\
        & SLTU$_{NET}$~\cite{zhang2023sltunet} & \checkmark & 52.92 &41.76 &33.99 &28.47 &52.11 \\
        \midrule
        \multirow{6}{*}{Gloss-free} 
        & CSGCR~\cite{zhao2021conditional}& \xmark & 36.71& 25.40& 18.86& 15.18 & -- \\
        & GFSLT-VLP~\cite{zhou2023gloss} & \checkmark & 43.71 &33.18 &26.11 &21.44 &42.29 \\
        & FLa-LLM~\cite{chen2024factorized}& \checkmark & 46.29& 35.33& 28.03& 23.09& 45.27 \\
        & Sign2GPT~\cite{wong2024sign2gpt}& \checkmark & 49.54 & 35.96 & 28.83& 22.52& \textbf{48.90} \\ 
        & SignLLM~\cite{gong2024llms} & \checkmark & 45.21 & 34.78 & 28.05 & 23.40 &44.49 \\ 
        
        & SpaMo~\cite{hwang-etal-2025-efficient} & \xmark & \underline{49.80}& \underline{37.32} &\underline{29.50} &\underline{24.32} & 46.57 \\
        \cdashline{2-8}
        & \textbf{SignMouth (Ours)}& \xmark & \textbf{50.37}& \textbf{37.75} &\textbf{29.94} &\textbf{24.71} &\underline{48.38} \\
        \bottomrule
        \end{tabular}
    }
    \caption{
    Comparison of model performance on the PHOENIX14T dataset. ``Vis. Fi'' specifies whether visual features are fine-tuned on sign language data. The best results are highlighted in bold, the second-best are \underline{underlined}, and “--” indicates missing or unavailable values. B1-B4 refer to BLEU-1 to BLEU-4, and RL refers to RougeL.
    }
    \label{tab:res_p14t_csl}
\end{table*}

\begin{table*}
\centering{
    \begin{tabular}{c l c c cccccc} \toprule
    \textbf{Setting}& \textbf{Methods} & \textbf{Vis. Fi}& \textbf{B1} &\textbf{B2} &\textbf{B3} &\textbf{B4} &\textbf{RL} \\ \midrule
    \multirow{2}{*}{Weakly Gloss-free}& GloFE-VN~\cite{lin2023gloss} & \(\checkmark\)& 14.94 &7.27 &3.93 &2.24 &12.61 \\
                                        & OpenSLT~\cite{tarres2023sign} & \(\checkmark\)& 34.01 &19.30 &12.18 &8.03 &- \\ \midrule
    \multirow{4}{*}{Gloss-free}& YT-ASL-SLT~\cite{uthus2024youtube} & \xmark& 14.96 & 5.11& 2.26& 1.22& -\\ 
                                & SSVP-SLT~\cite{rust2024towards} & \(\checkmark\)& 30.20& 16.70&  10.50& 7.00& 25.70& \\
                                & FLa-LLM~\cite{chen2024factorized} & \(\checkmark\)& 29.81& 18.99& 13.27& 9.66& 27.81\\
                                & SpaMo~\cite{hwang-etal-2025-efficient} & \xmark& \underline{33.41}& \underline{20.28}& \underline{13.96}& \underline{10.11}& \underline{30.56}&  \\ 
                                \cdashline{2-10}
                                & \textbf{SignMouth (Ours)} & \xmark& \textbf{33.88}& \textbf{21.04}& \textbf{14.23}& \textbf{10.75}& \textbf{31.44} \\ \bottomrule
    \end{tabular}
}
\caption{Comparison of model performance on the How2Sign dataset. ``Vis. Fi'' specifies whether visual features are fine-tuned on sign language data. The best results are highlighted in bold, the second-best are \underline{underlined}, and “--” indicates missing or unavailable values. B1-B4 refer to BLEU-1 to BLEU-4, and RL refers to RougeL.}
\label{tab:res_h2s}
\end{table*}

\subsection{Ablation Study}
\paragraph{Effect of different components}
Table~\ref{tab:component} presents an ablation study of key components in our framework. Using only the spatial encoder (SE) yields strong baseline performance (BLEU-4: 19.87, ROUGE: 41.38), while relying solely on the lip encoder (LE) results in significantly lower scores (BLEU-4: 12.73), indicating that mouthing features alone are insufficient.

Naively fusing SE and LE slightly degrades performance, highlighting the need for guided modality integration. Adding the Visual-Text Alignment (VT-Align) module to SE improves BLEU-4 to 20.72, demonstrating its effectiveness in bridging visual and textual semantics.

Combining SE and LE with VT-Align brings further improvements, confirming the complementarity of manual and non-manual features. The full model, integrating SE, LE, VT-Align, and Sign Modality Alignment (SM-Align), achieves the best results (BLEU-4: 24.71, ROUGE: 48.38), validating the effectiveness of our multi-level contrastive fusion strategy for gloss-free SLT.

\begin{table*}[t]
\small
\centering
\begin{tabular}{p{0.10\linewidth} p{0.85\linewidth}}
\toprule
\textbf{Input Type} & \textbf{Output (German + English Translation)} \\
\midrule
Spatial & am \textbf{montag} unbeständiges wetter stellenweise \textbf{auch} mal \textbf{sonne}. \textit{(On Monday, unsettled weather, occasionally some sun.)} \\
Mouthing & und nun die wettervorhersage für morgen mittwoch den neunten juli. \textit{(And now the weather forecast for tomorrow, Wednesday the ninth of July.)} \\
Fused & am \textbf{montag und dienstag wechselhaft hier und da zeigt sich auch} mal \textbf{die sonne}. \textit{(On Monday and Tuesday, changeable weather, here and there the sun also appears.)} \\
Target & montag und dienstag wechselhaft hier und da zeigt sich aber auch die sonne. \textit{(Monday and Tuesday, changeable weather, here and there the sun also appears.)} \\
\textit{Remark} & \textit{Fused model recovers temporal details (“Dienstag”) and improves fluency. Lip-only output is off-topic; spatial-only lacks completeness.} \\
\midrule
Spatial & \textbf{am nachmittag wird es} wieder \textbf{freundlicher}. \textit{(In the afternoon it becomes pleasant again.)} \\
Mouthing & ihnen einen schönen abend und machen sie es gut. \textit{(Have a nice evening and take care.)} \\
Fused & \textbf{am nachmittag wird es dann freundlicher}. \textit{(In the afternoon it then becomes pleasant.)} \\
Target & am nachmittag wird es dann freundlicher. \textit{(In the afternoon it then becomes pleasant.)} \\
\textit{Remark} & \textit{Fused result matches the target; “dann” is captured via lip cues.} \\
\midrule
Spatial & \textbf{heute nacht} elf \textbf{grad an der} ostsee und \textbf{vier grad} im erzgebirge. \textit{(Tonight, eleven degrees on the Baltic Sea and four degrees in the Ore Mountains.)} \\
Mouthing & \textbf{heute nacht} sechzehn \textbf{grad} im vogtland und vierzehn \textbf{grad} am oberrhein. \textit{(Tonight, sixteen degrees in Vogtland and fourteen degrees on the Upper Rhine.)} \\
Fused & \textbf{heute nacht zwölf grad an der nordsee vier grad in der eifel}. \textit{(Tonight, twelve degrees on the North Sea and four degrees in the Eifel.)} \\
Target & heute nacht zwölf grad an der nordsee vier grad in der eifel. \textit{(Tonight, twelve degrees on the North Sea and four degrees in the Eifel.)} \\
\textit{Remark} & \textit{Only fused model captures correct numerals and location—demonstrates strong grounding.} \\
\bottomrule
\end{tabular}
\caption{Qualitative examples comparing outputs from spatial-only, lip-only, and fused models. The fused model consistently improves semantic correctness and fluency. Correctly translated 1-grams are shown in bold.}
\label{tab:qualitative}
\end{table*}

\begin{table*}[ht]
\scriptsize
\centering
{%
    \renewcommand{\arraystretch}{0.95}
    \begin{tabular} {cccc  ccccc} \toprule
    \multicolumn{4}{c}{\textbf{Component}} & \multicolumn{5}{c}{\textbf{Metric}} \\ \cmidrule(lr){1-4} \cmidrule(lr){5-9}
    \textbf{SE}& \textbf{LE}& \textbf{VT-Align}& \textbf{SM-Align} & \textbf{B1}& \textbf{B2}& \textbf{B3}& \textbf{B4}& \textbf{RG} \\ \midrule 
    \(\checkmark\)& & & & 44.95& 32.04& 24.56& 19.87& 41.38 \\
    & \(\checkmark\)& & & 31.58& 23.34& 16.65& 12.73& 40.28 \\
    \(\checkmark\)& \(\checkmark\)& & & 43.90& 31.06& 23.59& 18.93& 40.53 \\
    \(\checkmark\)& & \(\checkmark\)& & 46.88& 34.59& 26.91& 20.72& 42.74 \\
    \(\checkmark\)&\(\checkmark\) & \(\checkmark\)& & 46.80& 34.26& 26.82& 22.33& 43.78 \\
    \cdashline{1-9}
    \(\checkmark\)& \(\checkmark\)& \(\checkmark\)& \(\checkmark\)&  \textbf{50.37}& \textbf{37.75} &\textbf{29.94} &\textbf{24.71} &\textbf{48.38} \\ \bottomrule
    \end{tabular}
}
\caption{Ablation study of the different components on
PHOENIX14T dataset. 
SE: Spatial encoder; 
LE: Lip-reading encoder;  
VT-Align: Visual-text alignment; 
SM-Align: Sign-mouthing alignment.}
\label{tab:component}
\end{table*}

\begin{table}[t]
\scriptsize
\centering{%
    \begin{tabular} {l c ccccc} \toprule
        \textbf{Models}& \textbf{Size}& \textbf{B1}& \textbf{B2}& \textbf{B3}& \textbf{B4}& \textbf{RG}\\ \midrule
        mT5-XL& 3.7B& 20.76 & 10.15 & 7.68 & 6.47 & 15.33\\
        Llama-3.2-3B& 3M& 33.87 & 20.03 & 13.52 & 9.92 & 27.59 \\ 
        Llama-3.1-8B& 8M&37.52& 23.47 & 16.58& 12.75 & 31.93\\ 
        \cdashline{1-7}
        Flan-T5-XL& 3B & \textbf{50.37}& \textbf{37.75} &\textbf{29.94} &\textbf{24.71} &\textbf{48.38} \\ \bottomrule
    \end{tabular}
}
\caption{Ablation study of the impact of LLM on PHOENIX14T dataset.}
\label{tab:llm}
\end{table}

\paragraph{Impact of LLM}
This experiment demonstrates that the selection of LLMs has a significant impact on downstream SLT tasks. 
Table~\ref{tab:llm} compares the performance of various models, including mT5-XL~\cite{xue2020mt5}, Llama-3.2-3B~\cite{dubey2024llama}, Llama-3.1-8B~\cite{dubey2024llama}, and Flan-T5-XL~\cite{chung2024scaling}. 
Compared to models trained solely with language modeling (e.g., mT5), those equipped with explicit instruction learning mechanisms, such as Flan-T5, are better suited for multimodal translation and achieve superior performance. Therefore, we adopt Flan-T5-XL as the default language decoder in this paper.

\begin{figure}[!t]
  \centering
  \includegraphics[width=1.0\linewidth]{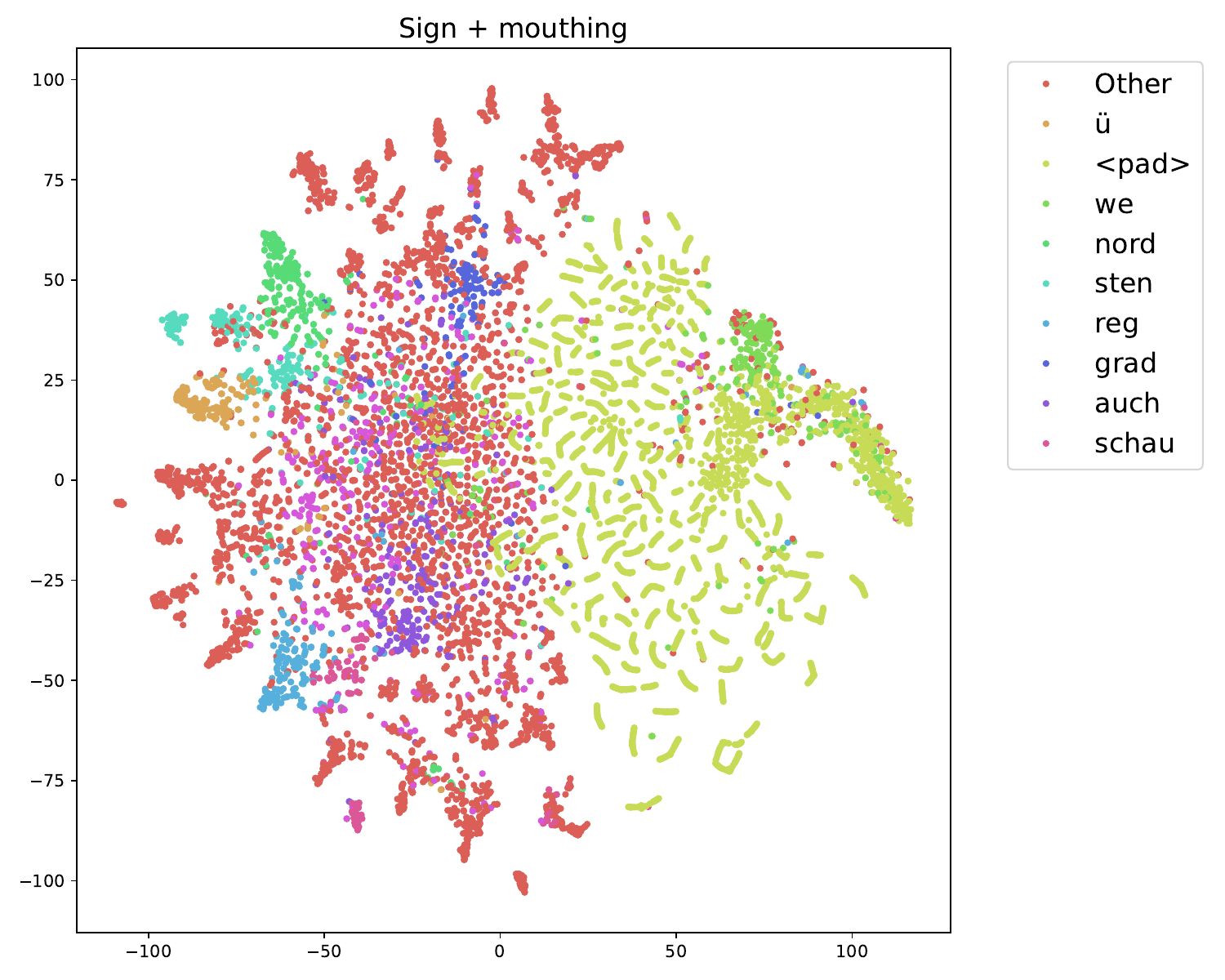} 
\caption{The t-SNE visualization of fused feature.}
\label{fig:vis}
\end{figure}

\subsection{Qualitative Analysis}
\paragraph{Qualitative translation examples}
We qualitatively compare the outputs from spatial-only, lip-only, and fused (spatial+lip) models in Table~\ref{tab:qualitative} on PHOENIX14T dataset. 
The results show that while spatial features capture core sign semantics, they sometimes overlook connective words and fine-grained details. 
In contrast, mouthing-only predictions often drift semantically, lacking grounding in manual gestures.
However, when combined, mouthing features provide valuable complementary cues—especially for functional words such as ``dann'' and temporal expressions like ``Montag und Dienstag'', which are crucial for sentence fluency and correctness.

These examples demonstrate that mouthing helps disambiguate signs and enhances temporal and syntactic understanding. 
The fused model produces more complete and accurate translations, particularly in capturing subtle but important elements such as numbers, days, or connectives. 
This validates the necessity of incorporating non-manual cues like mouthing to boost translation quality in gloss-free SLT systems.

\paragraph{Feature Visualization} To better understand the impact of gesture and mouthing fusion, we perform t-SNE visualization on the extracted features. As shown in Figure~\ref{fig:vis}, tokens like ``we'', ``nord'', and ``grad'' form distinct clusters in the joint embedding space, demonstrating that the model successfully captures meaningful semantic representations. 
The clear separation of non-linguistic tokens such as $<pad>$ and punctuation further validates the discriminative power of our multimodal fusion approach.

\section{Conclusion}
In this paper, we propose SignMouth, a novel framework to improve SLT accuracy. It fuses manual and non-manual cues, specifically spatial gesture and mouthing shape features. Besides, SignMouth introduces hierarchical contrastive learning with multi-level alignment objectives to ensure semantic consistency across sign-mouthing and visual-text modalities.
We conduct extensive experiments on benchmark datasets, PHOENIX14T and How2Sign, demonstrating that SignMouth achieves better performance in SLT. 

\bibliography{SignMouth}

\end{document}